\documentclass{article} 
\usepackage{nips14submit_e,times}

\pdfoutput=1

\usepackage{url}
\usepackage{amsmath,amsthm,amssymb,amsfonts,amsthm}
\usepackage{graphicx} 
\usepackage{wrapfig}
\usepackage{caption}
\usepackage{subcaption}

\usepackage{algorithm}    
\usepackage{algorithmic}
\usepackage{epstopdf}
\usepackage{soul}

\definecolor{lightgreen}{rgb}{.5,1,0}
\definecolor{lightred}{rgb}{10,0.2,0.2}
\definecolor{darkgreen}{rgb}{0,0.42,0}
\sethlcolor{lightgreen}

\newcommand{\myvec}[1]{\mathbf{#1}}
\newcommand{\myvecsym}[1]{\boldsymbol{#1}}

\newcommand{\vtheta}{\myvecsym{\theta}}

\newcommand{\vq}{\myvec{q}}

\newcommand{\vx}{\myvec{x}}
\newcommand{\be}{\begin{equation}}
\newcommand{\ee}{\end{equation}}
\newcommand{\bea}{\begin{eqnarray}}
\newcommand{\eea}{\end{eqnarray}}
\newcommand{\beaa}{\begin{eqnarray*}}
\newcommand{\eeaa}{\end{eqnarray*}}

\DeclareMathAlphabet{\mathpzc}{OT1}{pzc}{m}{n}
\title{Deep Multi-Instance Transfer Learning}

\author{
Dimitrios Kotzias$^{1,2}$ \quad Misha Denil$^2$ \quad Phil Blunsom$^{2,4}$ \quad Nando de Freitas$^{2,3,4}$  \\
$^1$University of California, Irvine\\
$^2$University of Oxford, United Kingdom\\
$^3$Canadian Institute for Advanced Research (CIFAR)\\
$^4$Google DeepMind\\
\texttt{dkotzias@ics.uci.edu}\\
\texttt{\{misha.denil,phil.blunsom,nando\}@cs.ox.ac.uk}
}

\nipsfinalcopy % Uncomment for camera-ready version

\begin{document}

\maketitle

\begin{abstract}
We present a new approach for transferring knowledge from groups to individuals that comprise them. We evaluate our method in text, by inferring the ratings of individual sentences using full-review ratings. This approach combines ideas from transfer learning, deep learning and multi-instance learning, and reduces the need for laborious human labelling of fine-grained data when abundant labels are available at the group level.  
\end{abstract}

%!TEX root = multi-instance deep learning.tex

\section{Introduction}

In many areas of human endeavour, such as marketing and voting, 
one encounters information at the group level. It might then be of interest
to infer information about specific individuals in the groups \cite{Kueck:2005}. 
As an illustrative example, assume we know the percentage of positive votes for each
neighbourhood of a city on a public policy issue. In addition, assume we have features for the individual voters.
This paper presents an approach for aggregating this
information to estimate the probability that a specific
individual, say Susan, voted positive. (If you're Susan, you should be
concerned about the privacy of your vote.)

This application is probably of questionable ethical value (other than as a warning on privacy issues), but
the same technology can be used to solve important problems
arising in artificial intelligence. 
In this work, we present a novel objective function, for instance learning in an a multi-instance learning setting~\cite{Dietterich:1997}. A similarity measure between instances is required in order to optimise the objective function. Deep Neural Networks have been very successful in creating representations of data, that capture their underlying characteristics~\cite{Hinton1986}. This work  capitalises on their success by using embeddings of data and their similarity, as produced by a deep network, as instances for experiments.

In  this paper we show that this idea can be used to infer ratings of sentences (individuals) from ratings of reviews (groups of sentences). This enables us to extract the most positive and negative sentences in a review. In applications where reviews are overwhelmingly positive, detecting negative comments is a key step toward improving costumer service. 
%Our approach also enables us to build sentiment analysis tools for reviews and sentiment-based search using review data. The latter application is important as the value of many products cannot be ascertained before consumption. Ideally, we would like to conduct queries such as: What's the best place for Chicken Tikka Masala in London?, or what is the best acting performances of Robert de Niro?
\begin{figure}[t]
\centering
\includegraphics[scale=.3]{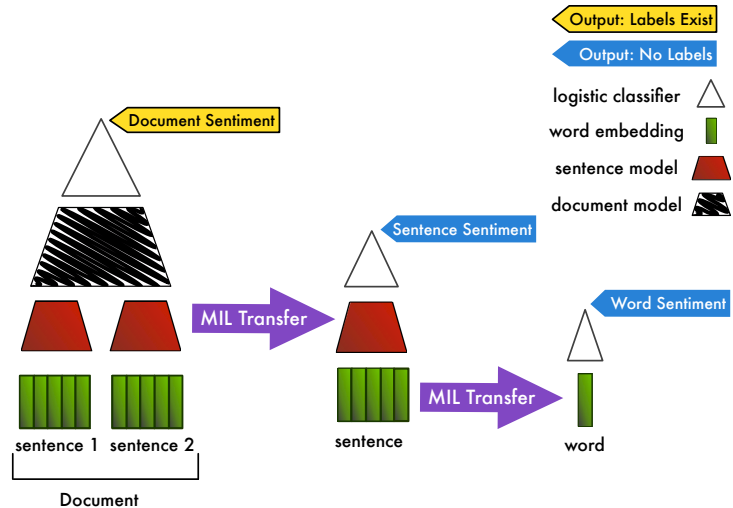}
\caption{Deep multi-instance transfer learning approach for review data.}
\label{fig:structure}
\end{figure}

Figure~\ref{fig:structure} presents an overview of our approach, which we refer to as deep multi-instance transfer learning. The first step in this approach involves creating a representation for sentences. We do that by training the supervised {document convolutional neural network} of Denil \emph{et al.}~\cite{Denil:2014} to predict review scores. 
As a result of this training, we obtain embeddings (vectors in a metric space) for words, sentences and reviews. These embeddings are the features for the individuals (sentences in this case). 
We chose this model, because it is the state of the art in convolutional networks, and the resulting sentence embeddings are not optimised for the problem we are attempting to solve. This adversarial scenario, illustrates the power of our model to work well, with multiple distributed representations of data. 

Using these features, we formulate a regularized manifold learning objective function to learn the labels of each sentence. That is, we transfer the labels from entire reviews to individual sentences and in doing so, we eliminate the high human cost of gathering labels for individual sentences.

%The embeddings enable our model to take advantage of the similarity between sentences. The objective function 
%to incur a cost if similar sentences have different costs. These two components, facilitate learning in a way that the score of each individual sentence, is affected by other similar sentences, even if they contain completely different words.

%!TEX root = multi-instance deep learning.tex

\section{Background}

\subsection{Deep Natural Language Processing}

Following the sweeping success of deep learning in computer vision, researchers in deep learning have begun to focus their efforts on other tasks.  In particular, applications of neural networks to natural language processing have received considerable attention.

Early work on applying neural networks to language models dates back several decades~\cite{Hinton1986,Bengio2003}.  More recently interest in this area has been renewed by the demonstration that many low level NLP tasks can be solved effectively with convolutional neural networks~\cite{Collobert:2011} and also by the development of training methods for distributed representations for words~\cite{Mikolov2013b,Mikolov2013a}.

Moving beyond representations for words, neural network models have also been used to build representations for larger blocks of text.  A notable example of this is the paragraph vector~\cite{Le2014} which extends the earlier work of Mikolov \emph{et al.}~\cite{Mikolov2013b} to simultaneously build representations for words and paragraphs.  Another recent move in this direction is the work of Denil \emph{et al.}~\cite{Denil:2014} which uses a convolutional neural network to build representations for words, sentences and documents simultaneously. We adopt this convolutional neural network for our experiments, however the proposed objective funtion is orthogonal to the method used to represent the data.

%!TEX root = multi-instance deep learning.tex

\subsection{Multi-Instance Learning}
Multi-instance Learning is a generalisation of supervised  learning, in which labels are associated with \emph{sets of instances}, often referred to as \emph{bags} or \emph{groups}, instead of individual instances. 
This powerful extension of supervised learning has been applied to a large variety of problems, including drug activity prediction \cite{Dietterich:1997}, content based image retrieval and classification \cite{Maron:1998,Yang:2000}, text categorization \cite{Andrews:2003,Andrews:2004}, music retrieval \cite{Mandel:2008}, translation and object recognition \cite{Kueck:2004,Carbonetto:2006b,Carbonetto:2008ijcv,Cour:2011} and privacy \cite{Kueck:2005,Kifer:2009}.  

While there are many variations of multi-instance learning, the key property is that each data example is a bag, and not a single individual. While the goal of some works is to predict labels for new groups, others focus on predicting labels for individual instances in the training or test datasets.

Prior work on Multi-instance learning differentiates in the assumptions made about the function that connects groups and instances. The initial formulation of the multi-instance learning problem by Dietterich \emph{et. al} \cite{Dietterich:1997} assumes that the label of each bag is binary, and that for a group to be positive, at least one the instances in the group must have a positive label. Weidmann  \emph{et. al} 
\cite{Weidmann:2003} consider a generalisation where the {presence} of a combination of instances determines the label of the bag. 
Xu \emph{et. al} \cite{Xu:2004} assume that all instances contribute equally and independently to a bag's class label, and the bag label is determined by the expected value of the population in the bag. In this work, we use this assumption to derive a regulariser that transfers label information from groups to individuals.

Recent works have considered generalizations where each bag is described in terms of the expected proportion of elements of each class within the bag. Here, the goal is to predict the label of each individual within the bags \cite{Kueck:2005,Quadrianto:2009}. For a recent survey on multi-instance learning, we refer the reader to \cite{Foulds:2010}. However, the literature on this topic is vast and that there is disagreement in the terminology. The closest works to ours are the ones of \cite{Kueck:2004,Kueck:2005,Quadrianto:2009,Liang:2009}.

%!TEX root = multi-instance deep learning.tex

\section{Deep Multi-Instance Transfer Learning}

In our formulation of deep multi-instance transfer learning, we are given a set of training instances 
\begin{align*}
\mathcal{X} = \{\vx_i\}_{i\in I}
\end{align*}
Unlike in the standard supervised setting, we are not given labels for each training instance directly.  Instead we are given labels assigned to groups of instances 
\begin{align*}
\mathcal{D} = \{(\mathcal{G}_g, s_g)\}_{g=1,\ldots,G}
\end{align*}
where $\mathcal{G}_g \subseteq \mathcal{X}$ is a mutli-set of instances from $\mathcal{X}$ and $s_g$ is a label assigned to the group $\mathcal{G}_g$, which we assume to be an unknown function of the (unobserved) labels of the elements of $\mathcal{G}_g$.  We are also equipped with a function $W(\vx_i, \vx_j) \in (0,1)$ which measures the similarity between pairs of instances. An example illustrating how we construct this similarity measure will be presented in the next section.

Our goals here are twofold.  Firstly, we would like to infer labels for each example by propagating information from the group labelling to the instances, essentially inverting the unknown label aggregation function on the training data.  To do this we take advantage of the similarity measure to compute a label assignment that is compatible with the group structure of the data, and that assigns the same label to similar instances.

Our second goal is more ambitious.  In addition to assigning labels to the training instances we also aim to produce a classifier $y(\vx, \vtheta)$ which is capable of predicting labels for instances not found in the training set.

%\note{We should compare to doing label assignment with MIL and then training a supervised classifier using the MIL labels}

We achieve both of these goals by constructing a training objective for the classifier $y(\vx,\vtheta)$ as follows:
\begin{align}
J(\vtheta) = 
\sum \limits_{i,j \in I} 
W(\vx_i,\vx_j) \left(
y(\vx_i,\vtheta) - y(\vx_j, \vtheta)
\right)^2
+ \lambda
\sum \limits_{g=1}^{G} 
\left(
\frac{1}{|\mathcal{G}_g|}\sum \limits_{i \in \mathcal{G}_g} y(\vx_i,\vtheta) 
- s_g
\right)^2
\label{eq:deep-mil}
\end{align}

%\note{We should try replacing squared error with cross entropy}

% \begin{itemize}
% \item $I$ is the set of all the items inside all the groups
% \item $\vx_i$ is the embedding of the sentence $i$
% \item $W(\vx_i,\vx_j)$ is a similarity measure between sentences $i, j$
% \item $y$ is any classifier parametrised by $\vtheta$
% \item $G$ is the set of all groups
% \item $s_g$ is the score for the group $g$
% \item $\lambda$ is a scalar, that acts as a relative balance between the two objectives.
% \end{itemize}

Both terms in this objective can be seen as different forms of label propagation.  The first term is a standard manifold-propagation term, which spreads label information over the data manifold in feature space.  A similar term often appears in semi-supervised learning problems, where the goal is to make predictions using a partially labelled data set.  In such a setting a label propagation term alone is sufficient; however, since we have labels only for groups of instances we require additional structure.

While we have adopted a weighted square-loss, any other weighted loss functions can be used as the first term of the objective function. It simply ensures that similar individual features $\vx_i$ are assigned similar labels $y$. 

The second term parametrises the whole-part relationship between the groups and the instances they contain, which has the effect of propagating information from the group labels to the instances.  Here we have chosen the simplest possible parametrisation of the whole-part relationship, which says that the label of a group is obtained by averaging the labels of its elements. 
This term acts as a regulariser and helps avoid the trivial cases where every instance has the same label, regardless of the group it belongs.  
%Naturally more complex whole-part relationships could be explored, but doing so is beyond the scope of this work. We have opted for simplicity and showing that this choice leads to very good experimental results.

Each individual term in the cost function by itself would not work well. This situation is not unlike what we find when we carry out kernel regression with $\ell_1$ regularization, where the likelihood term often leads to pathological problems and the regularizer simply has the effect of shrinking the parameters to a common value (typically zero). However, when we combine the two competing terms, we are able to obtain reasonable results.

%\note{What happens if we use only the second term?}

The parameter $\lambda$ trades off between the two terms in this objective.  The maximum theoretical value of the first term is $|I|^2$, since each summand falls in the interval $[0,1]$.  For the same reason, the second term is bounded by $|G|$. We therefore set $\lambda = \alpha\frac{|I|^2}{|G|}$ in order to trade off between their two contributions directly.  Of course it may not be the case that both terms are equally important for performance, which is why we have left $\alpha$ as a parameter.

Optimising this objective produces a classifier $y(\vx,\vtheta)$ which can assign labels to seen or unseen instances, despite having been trained using only group labels.  This classifier simultaneously achieves both of our stated goals: we can apply the classifier to instances of $\mathcal{X}$ in order to obtain labels for the training instances, and we can also use it to make predictions for unseen testing instances.

The power of this formulation relies on having a good similarity measure.  It would be simple to take the average score of each instance across groups, and minimise the second term of the objective.  However, the presence of the first term pushes similar items to have similar labels and allows for inter-group knowledge transfer.
%; in essence it allows for a softer classification of each item, depending on the complete dataset. 

%!TEX root = multi-instance deep learning.tex

\begin{figure}
\begin{center}
\includegraphics[width=0.6\linewidth]{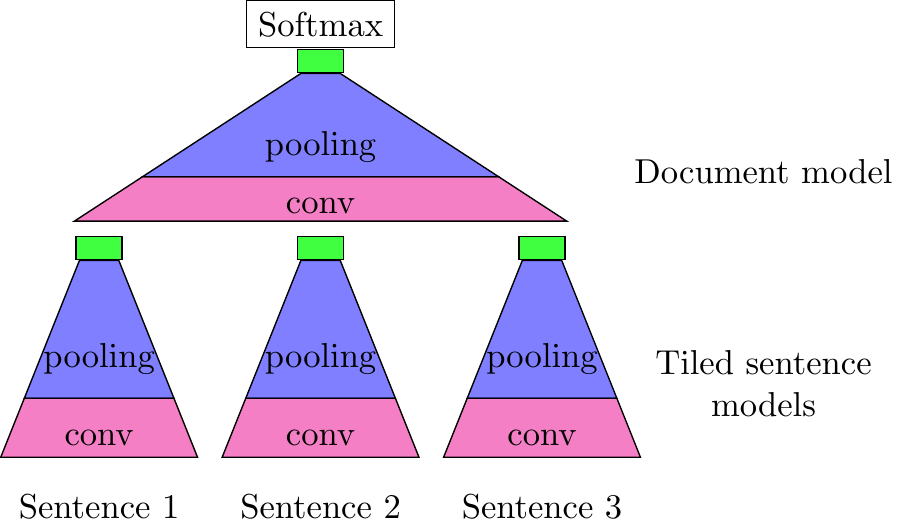}
\end{center}
\caption{Model from Denil~\textit{et~al.}~\cite{Denil:2014}.  The green squares indicate embedding vectors for sentences (atop the tiled sentence models) and for documents (atop the document model).  This model is ideal for our setting because it produces sentence embeddings as an intermediate representation but requires only document level labels for training.}
\label{fig:convnet}
\end{figure}

\section{From Review Sentiment to Sentence Sentiment}

\emph{Sentiment Attribution} refers to the problem of attributing the sentiment of a document to its sentences. Given a set of documents and a sentiment label for each one, we attempt to identify how much each sentence in each of these documents contributes positively or negatively towards its overall sentiment. This is a problem with an interesting set of applications, as it allows for a more efficient visualisation of documents, explores causality, and aids towards automatic summarisation.

We can express the sentiment attribution task as a deep multi instance learning learning problem by considering documents to be groups, and the sentences in each document to be individuals.  Following the procedure outlined in the previous section, we parametrise the relationship between sentence labels and document labels by assuming that the sentiment of a document is simply the average sentiment of its constituent sentences.

In order to obtain a similarity measure for sentences we take advantage of recent work in learning distributed representations for text.  Many works have shown how to capture the semantic relationships of words using the geometry of a continuous embedding space, and more recent works have extended this to learning representations of larger blocks of text~\cite{Le2014,Denil:2014}.

Given a distributed representation for each sentence in our training set we can create a similarity measure by setting
\begin{align*}
W(\vx_i, \vx_j) = \exp(-\|\vx_i - \vx_j\|_2^2)
\end{align*}
where $\vx$ represents the distributed representation of a sentence.
If the distributed representations have been created correctly then we should expect nearby points in embedding space to correspond to semantically similar sentences, making the Euclidian norm an appropriate measure of closeness.

We obtain sentence embeddings using the convolutional neural network from Denil et al.~\cite{Denil:2014}, which is particularly well matched to our setting.  This model is trained using only document level supervision but also produces word and sentence embeddings as an intermediate representation, which we can extract for our own purposes.  The structure of this model is shown in Figure~\ref{fig:convnet}.  We obtain these embeddings with a simple forward pass through the network, consider them instances and use the binary sentiment score of reviews as the group score and optimise our objective function with respect to our parameters $\vtheta$.

% This can be modelled as Deep Multi Instance Learning problem, by considering that the groups are the documents, and the sentences are the instances of each group. The score of the document, is based on a simple average on the score of its sentences. 
% The reason we model this as a Deep Multiple Instance Learning problem, is because we want to take advantage of the knowledge of the similarity of sentences. This is based on the belief that similar sentences, should have a similar sentiment score. This proves to be a good example of transfer learning, as the sentences fed to our model are embeddings  created by a single convolutional neural network. This network was optimised for a different task, however the embeddings still manage to provide the proper similarity information for this belief to hold true. 

\begin{figure}[t]
\small{\hl{Paul Bettany did a great role as the tortured father whose favorite little girl dies tragically of disease.\\
For that, he deserves all the credit.}\\
\textcolor{red}{However, the movie was mostly about exactly that, keeping the adventures of Darwin as he gathered data for his theories as incomplete stories told to children and skipping completely the disputes regarding his ideas.\\
Two things bothered me terribly: the soundtrack, with its whiny sound, practically shoving sadness down the throat of the viewer, and the movie trailer, showing some beautiful sceneries, the theological musings of him and his wife and the enthusiasm of his best friends as they prepare for a battle against blind faith, thus misrepresenting the movie completely.\\
To put it bluntly, if one were to remove the scenes of the movie trailer from the movie, the result would be a non descript family drama about a little child dying and the hardships of her parents as a result.}\\
\hl{
Clearly, not what I expected from a movie about Darwin, albeit the movie was beautifully interpreted.}}
\\[0.2cm]
\caption{For this review, our approach assigns positive sentiment to the first two and last sentences of the review. The remaining sentences are assigned negative sentiment.}
\label{fig:review_example}
\end{figure}

\subsection{Dataset and Experimental Setup}

For evaluating and exploring the problem of sentiment attribution, we use the IMDB movie review sentiment dataset originally introduced by Maas \emph{et~al.~}\cite{Maas:2011} as a benchmark for sentiment analysis. This dataset contains a total of 100,000 movie reviews posted on \texttt{imdb.com}. There are 50,000 unlabelled reviews and the remaining 50,000 are divided in a 25,000 review training set and a 25,000 review testing set. Each of the labelled reviews has a binary label, either positive or negative. In our experiments, we train only on the labelled part of the training set. 

We use NLTK\footnote{\url{http://www.nltk.org/}} to preprocess each review by first stripping the HTML markup, breaking it into sentences and then breaking each sentence into words. We also map numbers to a generic \texttt{NUMBER} token and any symbol that is not in \texttt{.?!} to \texttt{SYMBOL}. We replace all words that appear less than 5 times in the training set with \texttt{UNKNOWN}. This leaves us with a total of 29,493 words, 311,919 sentences in the training set and 305,929 sentences in the testing set.

We parametrise the model of Denil \textit{et al.}~\cite{Denil:2014} to obtain embeddings, $\vx_iin\mathbb{R}^{24}$, for sentences in the training and testing sets. This also results in word embeddings, which are not utilised in the score of this work. %. In this work, we do not attempt to transfer sentiment to the words.

For these experiments we used as our classifier a simple logistic regression, 
\[
y(\vx_i,\vtheta) = \sigma( \vtheta^\top\vx_i)= \frac{1}{1 + e^{-\vtheta^\top\vx_i}},
\] 
and set the regularisation coefficient in Equation~\ref{eq:deep-mil} to
$\lambda = 0.04 \frac{|I^2|}{|G|}$.

We optimize the objective function with stochastic gradient descent (SGD) for 1050 iterations with a learning rate of $\alpha = 0.0001$.
We used a mini-batch size of 50 documents, and carried out 7 SGD iterations in each mini-batch, for a total of 3 epochs. Different configurations showed very similar results to those reported.

The time required for training, is in the order of 3 minutes in a consumer laptop. Evaluation time is in the order of 0.1 seconds for all 305,929 sentences in the test set.

As a qualitative measure of the performance of our approach, Figure \ref{fig:review_example} illustrates the predicted sentiment for sentences in a review\footnote{\url{http://www.imdb.com/title/tt0974014/reviews}} from the test set. 
This is a particularly tricky example, as it contains both positive and negative sentences, which our model identifies correctly. Moreover, the largest part of this review is negative. Hence, the naive strategy of using a simple count of sentences to identify the total sentiment of  review, would fail in this example, which accompanied a rating of $8/10$. Our approach on the other hand enables us to extract sentences that best reflect the sentiment of the entire review, and score them at the same time. Averaging the predicted sentence scores correctly classifies this as a positive review.

%\begin{figure}[htb!]
%\centering
%\includegraphics[scale=.34]{artwork/eg2font}
%\caption{}
%\label{fig:review_example}
%\end{figure}
%

\subsection{Evaluation}

The purpose of our approach is to rely on supervision at the group level to obtain predictions for the individuals in the groups. This weak form of supervision is the most appealing feature of deep multi-instance transfer learning.

As a sanity check, we evaluate the performance of our model as a group (review) classifier. To accomplish this, we average the predicted scores for sentences in each review to classify the test and train set reviews as a whole.

The performance of the sentence score averaging classifier is comparable with the state-of-the art for review classification. The accuracy is 88.47\% on the test set and 94.21\% on the training set. We emphasize again, that the approach only has access to labels at the review level and must infer the labels of sentences even in the training set. The state-of-the-art on this data set is 92.58\% \cite{Le2014}. 

The good performance of our naive review classifier provides good indication that we have been able to transfer the review labels to infer labels for the sentences. Furthermore it is an indication that we have trained our classifier $y$ correctly.

\begin{table}[t]
\begin{center}
\begin{tabular}{|l|r|r|}
\hline
                   & Precision & Recall \\ \hline \hline
Socher~\emph{et al.}~\cite{Socher:2013}           & 84.5\%  & 100\%  \\ \hline
MIL Transfer & 85.5\%  & 100\%  \\ \hline
Socher~\emph{et al.}~\cite{Socher:2013} (ignoring neutral class)   & 84.7\%  & 76.2\% \\ \hline
MIL Transfer (ignoring neutral class)  & 92.6\% & 76.2\% \\ \hline
\end{tabular}
\end{center}
%\quad
%\begin{tabular}{|l|r|r|}
%\hline
%Sentences from Seen Data & Precision &Recall \\ \hline \hline
%Stanford Converted & 82.7\%  & 100\%  \\ \hline
%Stanford           & 82.58\%  & 77.5\% \\ \hline
%MIL Transfer  & 81.9\%  & 100\%  \\ \hline
%MIL Transfer b = 0.04 & 87.74\% & 77.5\% \\ \hline
%\end{tabular}
\caption{Sentence-level classification performance.}
\label{tab:sentence}
\end{table}

To further evaluate the sentence predictions, we manually labelled 2000 sentences from our dataset as either positive or negative\footnote{\url{https://www.cs.ox.ac.uk/people/phil.blunsom/handlabelled_sentence_sentiment.zip}}. We split this dataset in half, based on the split by Maas~\emph{et~al.}, and report the results of scoring sentences from the testing set.

We compared the performance of our approach on this dataset with the Sentiment Analysis tool described in Socher~\emph{et~al.}~\cite{Socher:2013}.  This tool is pre-trained and made available online through a web interface\footnote{\url{http://nlp.stanford.edu/sentiment/} Accessed 20th of June 2014} which we use predict labels for our test data.  It must be emphasized, that this method is trained with supervision at the phrase-level, while we only require supervision at the review level. It is expensive to obtain labels at the phrase-level, but there exist millions, perhaps billions, of labelled reviews online.

\captionsetup[subfigure]{labelfont=bf,justification=centering}

\clearpage
\begin{figure}
    \centering
    \begin{subfigure}[b]{0.32\textwidth}
    \centering
        \includegraphics[width=\textwidth]{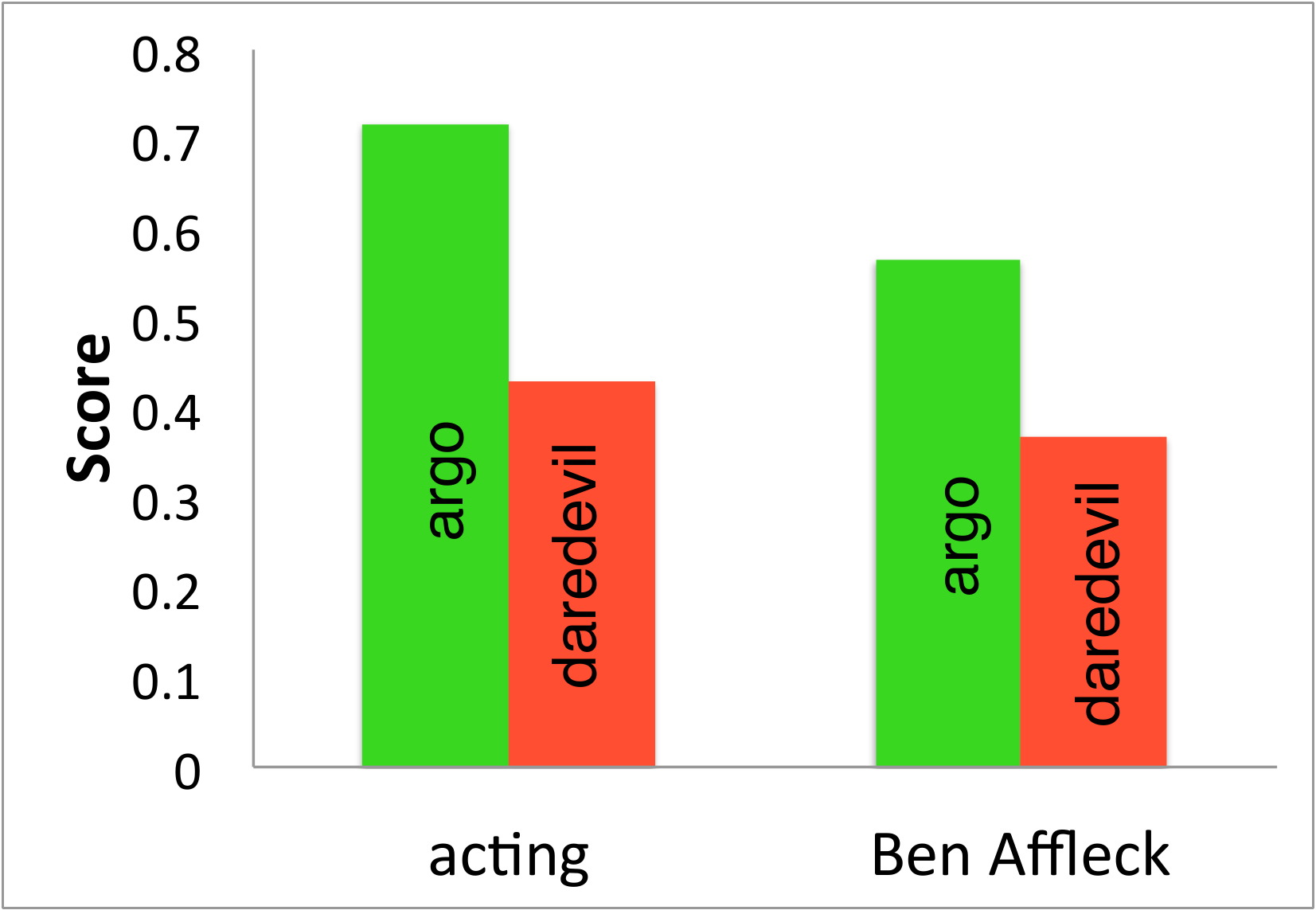}
        \caption{Ben Affleck \\\hspace{\textwidth} \emph{Argo, Daredevil}}
        \label{bef_affleck}
    \end{subfigure}%
    ~ 
    \begin{subfigure}[b]{0.32\textwidth}
    \centering
        \includegraphics[width=\textwidth]{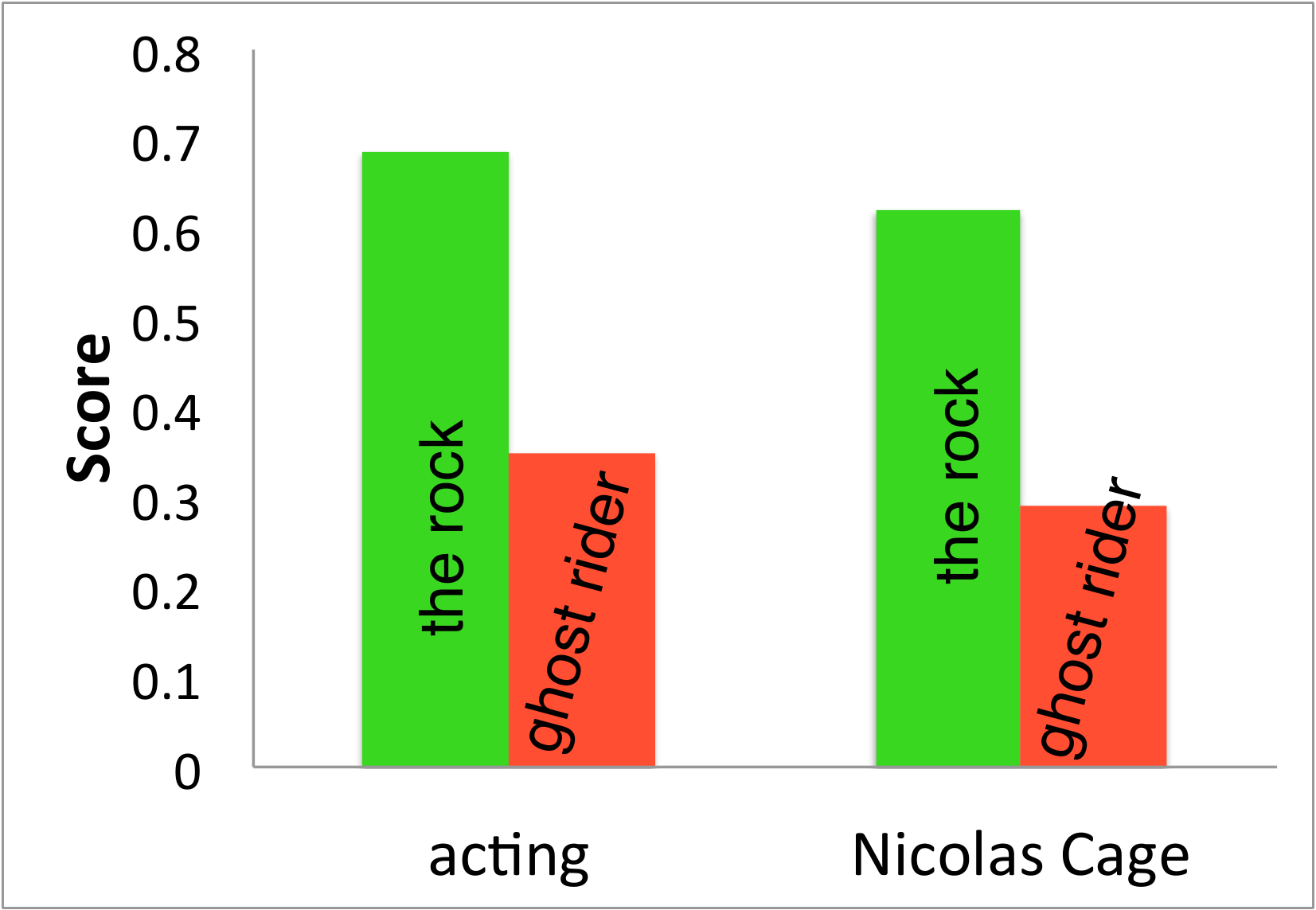}
        \caption{Nicolas Cage \\\hspace{\textwidth}\emph{The Rock, Ghost Rider}}
        \label{nicolas_cage}
    \end{subfigure}
    ~ 
    \begin{subfigure}[b]{0.32\textwidth}
    \centering
        \includegraphics[width=\textwidth]{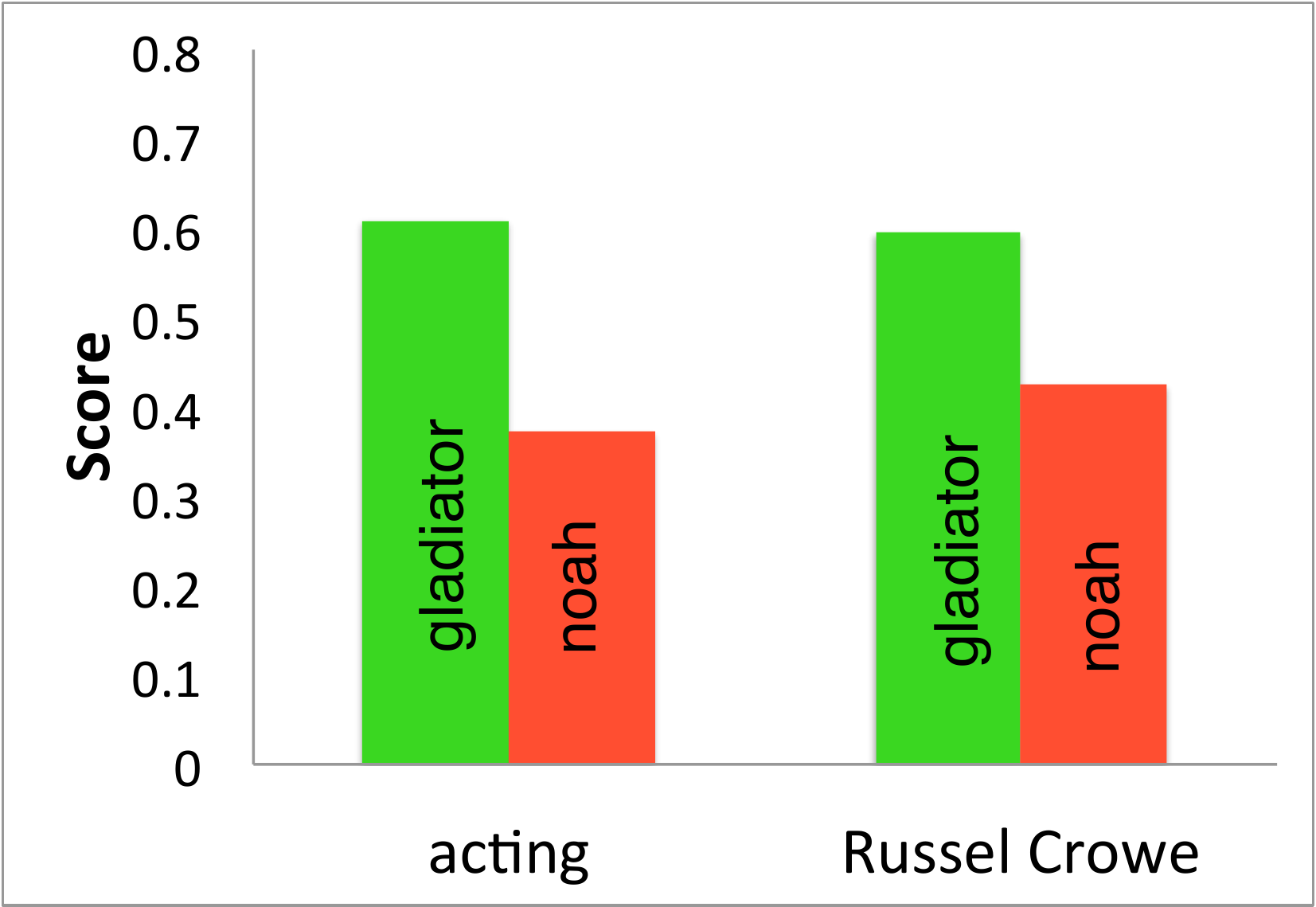}
        \caption{Russel Crowe \\\hspace{\textwidth} \emph{Gladiator, Noah}}
        \label{russel_crowe}
    \end{subfigure}
    \caption{Scores associated with the embedding of the word \textit{acting} and the protagonist names, when trained for different movies}\label{actors}
\end{figure}

\begin{figure}
\begin{center}
\includegraphics[width=1\linewidth]{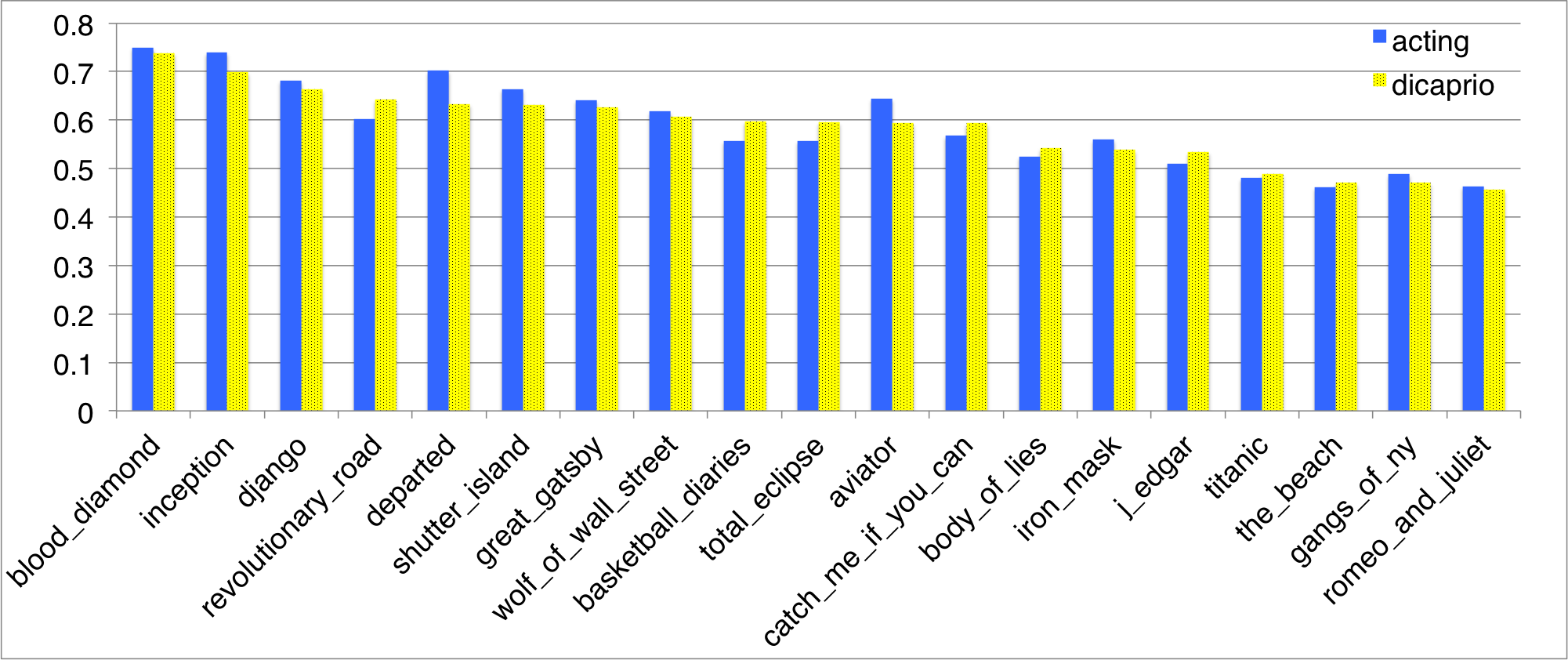}
\end{center}
\caption{Movies by actor \emph{Leonardo Di Caprio} sorted in order of the inferred sentiment for the embedding of his name, compared with the sentiment for the word \emph{acting}.}
\label{fig:dicaprio}
\end{figure}

\begin{figure}
\begin{center}
\includegraphics[width=1\linewidth]{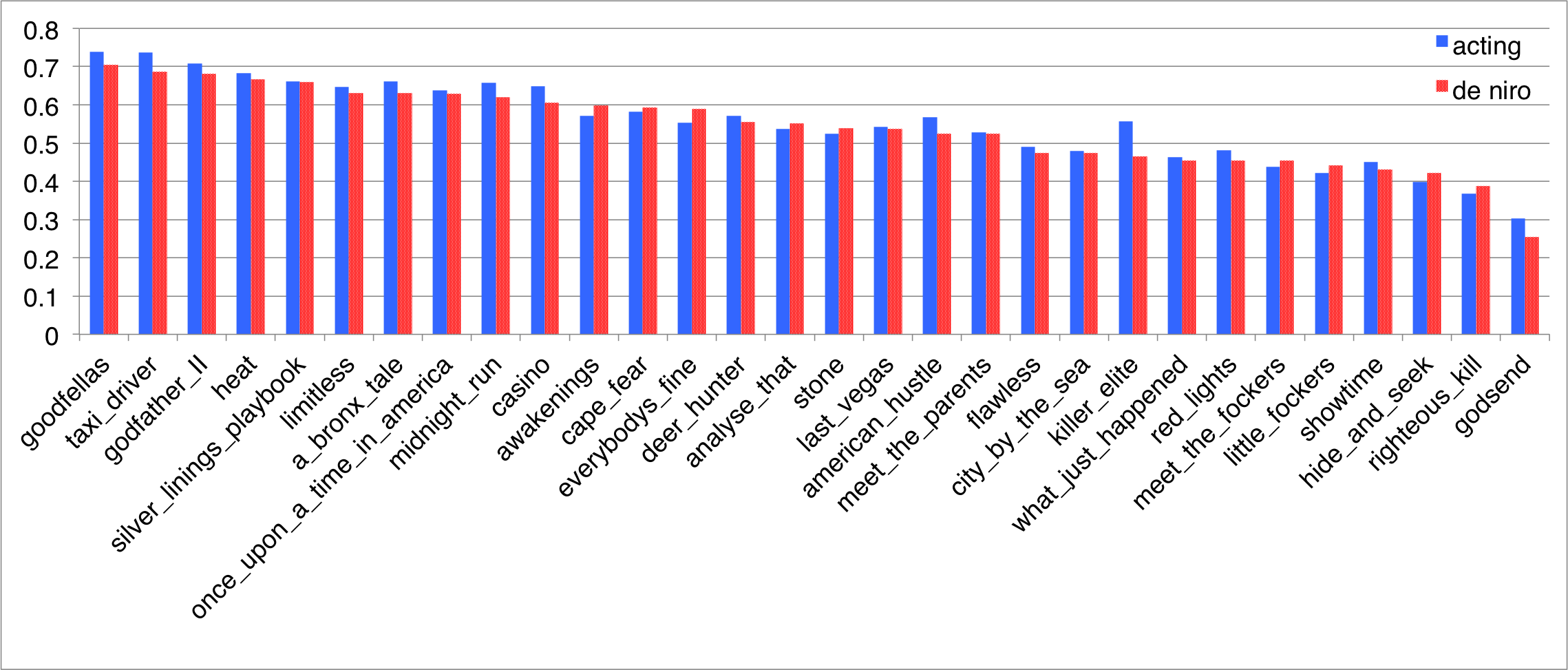}
\end{center}
\caption{Movies by actor \emph{Robert De Niro} sorted in order of the inferred sentiment for the embedding of his name, compared with the sentiment for the word \emph{acting}.}
\label{fig:deniro}
\end{figure}

\clearpage

The method of Socher~\emph{et al.}~\cite{Socher:2013} outputs the probability of a sentence belonging to the following five classes: \texttt{[Very Negative, Negative, Neutral, Positive, Very Positive]}. Subsequently, it chooses the class of highest probability as the predicted class. 
To convert this output to a binary decision, we count both \texttt{Positive} and \texttt{Very Positive} labels as positive, and do the same for negative labels. To manage the \texttt{Neutral} class, we consider two strategies. First, we ignore sentences for which the prediction is \texttt{Neutral} in the test set, which has the effect of reducing recall. Second, when the label of highest probability is \texttt{Neutral}, we use the label of second highest probability to decide whether the review is positive or negative. We report results using both scoring strategies.  As shown in Table~\ref{tab:sentence}, both strategies achieve similar precision.

Table~\ref{tab:sentence} also shows that our deep multi-instance transfer learning approach achieves higher precision for 100\% recall. In order to generate a neutral class with our approach, we introduce a boundary threshold $b$ and label sentences whose score  falls in the range $( 0.5-b,0.5 +b)$ as \texttt{Neutral}. We set $b= 0.048$ to calibrate to the recall level as Socher~\emph{et al.}~\cite{Socher:2013} when sentences predicted as \texttt{Neutral} are ignored. For the same recall, deep multi-instance learning obtains much higher precision.

In spite of the fact that deep multi-instance transfer learning requires much less supervision, it is able to obtain better sentiment predictions for sentences than a state-of-the-art supervised learning approach.

Finally we show how our multi-instance learning approach can be used to obtain entity level sentiment in a specific context.  For example, we can predict the sentiment associated with a particular entity (e.g., Leonardo di Caprio) in a chosen context (e.g. a movie).  To accomplish this we restrict our training data reviews of the chosen movie, and train a multi-instance classifier on this restricted data.  This restriction forces the model to predict sentiment within a specific context. After getting the representation of the sentence in metric space $\vq$, we can use the context-specific classifier $\vtheta_c$, to predict the sentiment associated with it, $y(\vq,\vtheta_c)$. If the phrase is an actor's name, we essentially obtain sentiment about his role in a specific movie.

Figure \ref{actors} illustrates the scores that the same actor achieved in two different movies. The total imdb movie scores agree with the ranking at each case, but more importantly this indicates how the same phrase, can have a completely different sentiment in a different context, which is desirable when ranking queries.

Figures~\ref{fig:dicaprio} and~\ref{fig:deniro} show this for a series of movies with the actors Leonardo di Caprio and Robert de Niro as the protagonist. The rankings are sorted based on the performance of the actor, and appear to be reasonable thus providing a visual indication that the approach is working well.

\section{Concluding Remarks}

This work capitalises on the advances and success of deep learning to create a model that considers similarity between embeddings to solve the multi-instance learning problem. In addition, it demonstrates the value of transferring embeddings learned in deep models to reduce the problem of having to label individual data items when group labels are available. Future work will focus on exploring different choices of classifiers, embedding models, other data modalities, as well as further development of applications of this idea.

\small{
%\subsubsection*{References}
\bibliography{deepBib}
\bibliographystyle{plainabbrv}
}

\appendix
\clearpage

\end{document}